\documentclass[a4paper]{jpconf}
\usepackage{graphicx}
\usepackage{amsmath}
\usepackage{amsfonts}
\usepackage{hyperref}
\usepackage{xfrac}
\usepackage{verbatim}
\usepackage{subfig}
\begin{document}
\title{Multivariate regression and fit function uncertainty}

\author{P\'eter K\"oves\'arki$^{1}$, Ian C Brock$^{2}$}

\address{${}^{1}$University of Wroclaw, Poland;${}^{2}$University of Bonn, Germany}

\ead{Peter.Koevesarki@cern.ch}

\newcommand{\prob}[1]{\mathrm{P}( #1 )}
\newcommand{\avg}[1]{\langle #1 \rangle}
\newcommand{\cov}[1]{\text{Cov}\left( #1 \right)}

\begin{abstract}
This article describes a multivariate polynomial regression method where the uncertainty of the 
input parameters are approximated with Gaussian distributions, derived from the central limit 
theorem for large weighted sums, directly from the training sample. The estimated uncertainties 
can be propagated into the optimal fit function, as an alternative to the statistical bootstrap 
method. This uncertainty can be propagated further into a loss function like quantity, with which it 
is possible to calculate the expected loss function, and allows to select the optimal polynomial 
degree with statistical significance. Combined with simple phase space splitting methods, it is 
possible to model most features of the training data even with low degree polynomials or constants.

\end{abstract}

\section{Introduction}

Regression methods are frequently used in particle physics, usually to quantify a continuous curve or 
surface that simplifies statistical sample. Typical examples are the calibration curves for certain 
detector responses and neural networks trained to identify particle. The mathematical goal is finding 
a $f:x\rightarrow y$ map between the $x$ input space to the $y$ target space in such a way that 
$f(x)$ predicts the $\mathbb{E}(y|x)$ conditional expectation value with statistical certainty. The least 
squares algorithm is known to converge to the conditional mean, given it is a finite number and the 
parametric $f$ function is in the family that contains the solution. This latter information is not always 
given and one must chose a function family general enough to cover unexpected features. Such a 
function family are the logistic functions and the radial base functions and generally the \emph{kernels}. For 
these one usually has to determine an ideal degree of freedom for the fit, namely the number of base 
functions to be used in order to avoid overtraining of the data and so avoiding picking up non-significant 
features from the statistical fluctuation. Although these are straightforward procedures, it is 
computationally intensive to find the global minimum of the sum of squares for the fit. A usually 
unexploited feature of the least squares method is that the global minimum can be exactly 
determined for kernels with fixed position in the $x$ space, because the amplitude that minimizes 
the sum of squares can be calculated with a linear equation, without numerical optimization.

With given $k_i(x), i\in{1..n_d}$ kernels and $a_i$ amplitudes the sum of squares for the data points $\{x_j, y_j\}, j \in \{1..N\} $ will take the form

\begin{equation}\label{eq:LeastSquareAmplitudeFit}
 E_{\chi^2} = \frac{1}{N}  \sum_j \left(  y_j - \sum_i a_i k_i\left( x_j\right)  \right)^2 = 
	\avg{y^2} - \sum_i 2a_i\avg{yk_i(x)} + \sum_{il} a_ia_l\avg{k_i(x)k_l(x)} \,,
\end{equation} 
where the angled brackets $\avg{ }$ indicate averaging over the sample. The \emph{loss function} $E_{\chi^2}$ in eq.~\eqref{eq:LeastSquareAmplitudeFit} can be minimised in respect of the $f_i$ amplitudes with 
\begin{equation}\label{eq:OptimalAmplitudes}
f_i = \sum_l G_{il}^{-1} h_l
\end{equation}
by using the matrix $G_{il} = \avg{k_i(x)k_l(x)}$ and the vector $h_i = \avg{yk_i}$. The $G_{il}$ matrix is symmetric and has $\frac{1}{2}n_d(n_d-1)$ parameters. A possible way to decrease the number of parameters is to use kernels which are power series $k_i(x) = k_1^i(x)$, resulting in a Hankel-type matrix $G_{il} = \avg{k_1(x)^{i+l}}$. A simple power series kernel might be based on the monomials, $k_1(x) = x, k_i(x) = x^i$, resulting in polynomial fitting. An other advantage of fitting a fix degree polynomial instead of Gaussian or a sigmoid kernels is that polynomials are not sensitive to the shift of features in the data, in other words they are translation invariant.

\section{Uncertainty and covariance of large weighted random sums}

The advantage of using matrix formalism in eq.~\eqref{eq:LeastSquareAmplitudeFit} is that the 
original training data $\{x_j, y_j\}$ is compressed into the $h_i$  vector and the $G_{il}$ matrix, which 
can be a great reduction in the number of input parameters. These input parameters are themselves 
random variables having a certain distribution that in principle could be derived from the generating 
distribution of the $\{x_j, y_j\}$ sample and the number of measurements $N$. Due to the central 
limit theorem, the generating distribution itself is not need to be known, only that it fulfils certain 
criteria, since $h_i$ and $G_{il}$ are generated by summing up random variables. Probably the most 
widely known of the central limit theorems is the one stating that if the generating distribution of the 
$X_i, i\in\{1..N\}$ random variables has finite mean $\bar X$ and variance $\sigma_X$, then the distribution 
of the variable $s = \frac{1}{\sigma_X}(\sum_i X_i/N - \bar X) $ converges to the normal distribution. 

The approximation of the covariance matrix of the $p_m = (h_1,...,h_{n_d}, g_1, ..., g_{2{n_d}}), m
\in \{1,...,3{n_d}\}$ input parameters for polynomial regression is the following. In a general formalism, every data point with index $j$ are triplets, consisting of an input value $x_j$, a target value $y_j$ and a weight $w_j$. With $b \in \{0,1\}$, $k_0 \{1,..., {n_d}\}$, $k_1\in\{1,...,2{n_d}\}$ the input parameter with pseudo index  $m ={k_b + b{n_d}}$ is calculated as

$$ p_m = \frac{1}{\sum_j w_j}\sum_j w_j y_j^b x_j^{k_b}\,.$$
The product $a_j = w_j y_j^b x_j^{k_b}$ for each $i$ index can be treated as a compound random variable. The $\sfrac{1}{\sum_j w_j}$ was not introduced into this new variable, because that would ruin the independence. Let\rq{}s define the following variables in order to estimate the probability distribution of $p_m$ . Let $\avg{aw} = \frac{1}{N} \sum_j w_j a_j$ be the weighted average of $a_j$. This is approximately a Gaussian variable with mean $\avg{aw}$ and variance $\sigma^2_{\avg{aw}} = \frac{1}{N(N-1)}\sum_j{(a_j w_j - \avg{aw})^2}$. Define the average weight similarly as $\avg{w} = \frac{1}{N}\sum_j w_j$, which is also distributed as a Gaussian with mean $\avg{w}$ and variance $\sigma^2_{\avg{w}} = \frac{1}{N(N-1)}\sum_j{( w_j - \avg{w})^2}$. These variables are indeed correlated, and their covariance is $\cov{\avg{aw}, \avg{w}} = \frac{1}{N(N-1)}\sum_j{(a_j w_j - \avg{aw})( w_j - \avg{w}) }$. The above definitions show, that $p_m$ is a ratio of two Gaussian variables. Its expectation value is $p_m$, while its variance can be approximated with error propagation. Assuming that $\sigma_{\avg{w}}/\avg{w} \gg 0 $, the approximation of the variance of $p_m$ is 

\begin{align}
 \sigma_{p_m}^2 &= \begin{pmatrix} \frac{\partial p_m}{\partial \avg{aw}}  \\[0.3em] \frac{\partial p_m}{\partial \avg{w}} \end{pmatrix}^\text{T}  \begin{pmatrix}   \sigma^2_{\avg{aw}} & \cov{\avg{aw}, \avg{w}} \\ \cov{\avg{aw}, \avg{w}} & \sigma^2_{\avg{w}}\end{pmatrix}  \begin{pmatrix} \frac{\partial p_m}{\partial \avg{aw}}  \\[0.3em] \frac{\partial p_m}{\partial \avg{w}} \end{pmatrix}  \nonumber\\
	&= \dfrac{1}{\avg{w}^2} \sigma_{\avg{aw}}^2  + \dfrac{\avg{aw}^2}{\avg{w}^4} \sigma_{\avg{w}}^2 - 2\dfrac{\avg{aw}}{\avg{w}^3}\cov{\avg{aw}, \avg{w}} \label{eq:WeightedSigmaFull} \\
	& = \dfrac{1}{\avg{w}^2} \sum_j \dfrac{w_j^2(\avg{aw} - a_j)^2}{N(N-1)} \,. \label{eq:WeightedSigmaShort}
\end{align} 
It can be seen that eq.~\eqref{eq:WeightedSigmaShort} gives back the known formula for the standard deviation of $\avg{a}$ when all weights are $w_j = 1$, which is also true when the weights are independent of the distribution of $a$. Similar derivation shows that the covariance between variables $p_{m_1}$ and $p_{m_2}$, with pseudoindices $m_1 ={k_{b_1} + b_1{n_d}}$ and $m_2 ={k_{b_2} + b_2{n_d}}$ can be calculated as 

\begin{equation}\label{eq:WeightedCovariance}
 \cov{p_{m_1}, p_{m_2}} =  \dfrac{1}{\avg{w}^2} \sum_j \dfrac{w_j^2(\avg{wy^{b_1}x^{k_{b_1}}} - y_j^{b_1} x_j^{k_{b_1}})(\avg{wy^{b_2}x^{k_{b_2}}} - y_j^{b_2} x_j^{k_{b_2}})}{N(N-1)} \,.
\end{equation}
 
%
%
%

The covariance and the variance estimation can be generalized to non-monomial kernels, by replacing $x^k$ in the above equations with the given kernel.

It must be noted that traditionally the uncertainty estimates on the fit parameters have different formulas. In most cases the sample $\{x_j, y_j\}$ is augmented with the uncertainty of the target, the conditional variance in the $y$ direction, $\sigma_{y_j}^2$, which can be considered as prior knowledge. In that case the data points in the formation of the estimation of expectation values receive a $\sigma_{y_j}^{-2}$ weight and a $c = \sum_j \sigma_{y_j}^{-2}$ normalization factor. This choice of weight comes from the principle that  the different measurements should be combined in a way that minimizes the uncertainty of the result, in this case the estimation of the expectation values. An example could be the $F(x) = a+bx$ least squares regression with $a,b$ unknowns on the $\{x_j, y_j, \sigma_{y_j}^2\}$ sample. The input parameters to this fit are 

$$ c = \sum_j \frac{1}{\sigma_{y_j}^{2}} \,,  h_0 = \avg{y} = \frac{1}{c} \sum_j \frac{y_j}{\sigma_{y_j}^2}\,, h_1 = \avg{yx} = \frac{1}{c} \sum_j \frac{y_jx_j}{\sigma_{y_j}^2}$$
$$ g_0 = \avg{x^0} = 1 \,, g_1 = \avg{x} = \frac{1}{c} \sum_j \frac{x_j}{\sigma_{y_j}^2}\,,
 g_2 = \avg{x^2} = \frac{1}{c} \sum_j \frac{x_j^2}{\sigma_{y_j}^2}\,.$$
 
 The optimal fit parameters are 
 
 $$ \begin{pmatrix} a \\ b \end{pmatrix} = \begin{pmatrix} g_0 & g_1 \\ g_1 & g_2 \end{pmatrix}^{-1}\begin{pmatrix} h_0 \\ h_1 \end{pmatrix} $$
As $a$ and $b$ are linear functions of $y_j$, it is easy to calculate the expectation values needed for the covariance matrix:

\begin{equation}\label{eq:CovarianceYuncertainty}
 \cov{a,b} = \mathbb{E}(ab) - \mathbb{E}(a)\mathbb{E}(b) = \frac{1}{c}\sum_j \frac{\partial a}{\partial y_j} \frac{\partial a}{\partial y_j} \sigma_{y_i}^2  = \avg{x} = g_1 \,.
 \end{equation}
Similarly, $\cov{a,a} = \sigma_a^2 = g_0$ and $\cov{b,b} = \sigma_b^2 = g_2$. This covariance matrix 
indeed differ from the one derived in eq.~\eqref{eq:WeightedCovariance}. The origin of the difference 
relies in the prior information that was built into the equations. In the case of eq.~\eqref{eq:WeightedCovariance}
the weights were provided with the data points, while in the case of eq.~\eqref{eq:CovarianceYuncertainty}
the $\sigma_{y_i}^2$ was given -- knowledge of the uncertainty on the $\mathbb{E}(y|x)$ conditional 
mean. The weights in the former case may come from Monte Carlo integration techniques or from 
weighted sample separation and it is thought to be fundamentally fixed, while in the latter case it is 
derived from the principle of optimal data combination. Though it is unclear whether the two methods 
could be combined, but the former method is thought to be superior as it was designed to approximate 
the $\sigma_{y(x)}^{2}$ from the sample itself, and also takes into account the uncertainty in the sampling 
of the input space $x$. Furthermore, it handles negative and  zero weights correctly.

\section{Fit function uncertainty}

With the knowledge of the uncertainty of the input parameters  $p_m = (h_1,...,h_{n_d}, g_1, ..., g_{2{n_d}}), m \in \{1,...,3{n_d}\}$, one can estimate the uncertainty of the fitted kernel amplitudes using linear error propagation in eq.~\eqref{eq:OptimalAmplitudes}. The first derivatives of $f_j = \sum_l G_{il}^{-1}h_l, G_{il} = g_{i+l}$ are

$$ \frac{\partial f_i}{\partial \avg{h_l}} = G_{il}^{-1} $$
$$ \frac{\partial f_i}{\partial \avg{g_o}} = -\sum_{lmn} G_{im}^{-1} \frac{\partial G_{mn} }{\partial{ g_o}} G_{nl}^{-1}h_{l}  \,.$$
Together with the previously calculated covariace matrix, the uncertainty of the fit function $ F(x) = \sum_i f_i x^i  $ at a given $x$ point is
$$ \sigma^2_{F(x)} =\sum_{nm} \frac{\partial \sum_i f_i x^i }{\partial p_m} \cov{p_m, p_n}  \frac{\partial \sum_l f_l x^l }{\partial p_n}  \,.$$

The uncertainty of the fitted function does not necessarily cover the true $\mathbb{E}(y|x)$ conditional mean.
 That only happens if the fit function is general enough to discover all the features of the sample. The
 meaning of this uncertainty is deeply routed in the central limit theorem. When the central limit theorem was
 applied to the $p_m$ input parameters, only the fact that the distribution of certain sums can be modeled
with a Gaussian distribution came from the theorem, the width and the mean of this Gaussian came from a maximum   
likelihood fit. This is typically interpreted as a posterior distribution for the true mean, but it can also be
 interpreted as a model fitted to the sample and predicting where the sum may converge with additional data
 points. The same can be said about the Gaussian uncertainty of the fit function. It tells us the likelihood
 where the fit function with the same degrees of freedom would converge with additional data, and not the position of
 the conditional mean. This is why some methodology is needed to compare fit functions with different degrees
 of freedom and see if the sample can be described better with one or the oter.

\section{The uncertainty of the loss function}

In the case of polynomial fitting one has to determine the degree of polynomial that is still statistically
 meaningful. Using too many degrees of freedom in a fit can result in overfitting or eventually in
 interpolation of the data points. In the latter case the $ E_{\chi^2} $ loss function simply reaches its
 absolute minimum, zero. However, the uncertainty of the fitted function increases with the number of degrees
 of freedom and this can be exploited in order to select significant features only, though one has to keep in
 mind that the Gaussian approximation of the distribution of the $p_m$ input parameters has a limitation.
 First, the Gaussian approximation is only true if the number of input points $N$ is large enough. Second, the
 uncertainty of the estimated covariance matrix may also increase to be comparable with the covariance matrix
 itself if the number of degrees of freedom in the fit is comparable to the number of sample points.

The na\"ive way of comparing the optimized $F^\text{opt}_{d_1}(x) =  \sum_{ik}^{d_1} h_i G_{d_1,ik}^{-1} x^k $ with degrees of freedom $d_1$ to $F^\text{opt}_{d_2}(x)$ with degrees of freedom $d_2$ would be calculating their loss functions $E_{d_1,\chi^2} = \frac{1}{N}\sum_j (y_i - F^\text{opt}_{d_1}(x))^2 $ and $E_{d_2,\chi^2}$ and checking whether their difference is significantly different then zero. This procedure does not work, as the approximate distribution of $E_{d_1,\chi^2}$ is not a good measure of fitness after the $F^\text{opt}_{d_1}(x)$ is substituted. It can be seen on an example where the $x$ space is thought to be non-random and only the $y$ coordinates of the sample points might vary when a new sample is obtained. After the substitution of 
$F^\text{opt}_{d}(x)$ into $E_{d,\chi^2}$ it can be simplified to 
\begin{align}
E_{d,\chi^2}  & =  \avg{y^2} - 2\sum_i F^\text{opt}_{d,i} h_i +\sum_{ik} F^\text{opt}_{d,i} G_{ik} F^\text{opt}_{d,k} \nonumber\\
	& = \avg{y^2} -  \sum_{ik} h_i G_{ik}^{-1} h_j\,,   \nonumber
\end{align}
which has an approximate expectation value 
\begin{align}
 \mathbb{E}(E_{d,\chi^2} ) & = \avg{y^2} - \mathbb{E}( \sum_{ik} h_i G_{ik}^{-1} h_j  ) \nonumber\\
 & =   \avg{y^2} -   \sum_{ik} h_i G_{ik}^{-1} h_j -  \sum_{ik} G_{ik} \cov{h_i, h_k}\,.  \nonumber
\end{align}
It contains a bias term compared to $E_{d,\chi^2}$ and allows $\mathbb{E}(E_{d,\chi^2} )$ to be lower than $E_{d,\chi^2}$. This bias means that the approximate distribution of $E_{d,\chi^2}$ is not related to the goodness of fit anymore, but to the possible $E_{d,\chi^2}$ minima. 

Ideally, the best measure of fitness would be a distance-like variable between the fit function and the real
 $\mathbb{E}(y|x)$ conditional expectation value. A good approximation to that is of course the loss function
  applied to $F^\text{opt}_{d}(x)$, but one must differentiate it from  the previously described $E_{d,\chi^2}$. In this picture, one must treat the sample as an approximation to $\mathbb{E}(y|x)$, and not as a
random variable. Let's call it the cross validation loss function, where only the $F^\text{opt}_{d}(x)$ is a
 random variable:

$$ E_{d,\chi^2}^+  =  \underbrace{\avg{y^2}}_{\text{fixed}} - 2\sum_i \underbrace{F^\text{opt}_{d,i}}
_{\text{varied}} \underbrace{h_i}_{\text{fixed}} + \sum_{ik} \underbrace{F^\text{opt}_{d,i}}_{\text{varied}} 
\underbrace{G_{ik} }_{\text{fixed}}\underbrace{F^\text{opt}_{d,k}}_{\text{varied}}\,.$$
As the first derivative of $\frac{\partial E_{d,\chi^2}^+}{\partial p_m' } = 0$ at $p_m' = \mathbb{E}(p_m)$, the expectation value of $E_{d,\chi^2}^+$ can be approximated through its second derivative as
\begin{align}
 \mathbb{E}(E_{d,\chi^2}^+ ) &= \avg{y^2} - 2\sum_i h_i \mathbb{E}(F^\text{opt}_{d,i}) + \sum_{ik}G_{ik}  
 \mathbb{E}(F^\text{opt}_{d,i} F^\text{opt}_{d,k}) \nonumber\\
\label{eq:ExpectedLossFunction} & = E_{d,\chi^2}^+ + \frac{1}{2}\sum_{lm} \frac{\partial^2 E_{d,\chi^2}^+}{\partial p_m \partial p_l} \cov{p_m, p_l} \,,
\end{align}
where the second derivative of $E_{d,\chi^2}^+$ was taken at the expectation values of $p_m$, and can be expressed as a block matrix
$$ \frac{\partial^2 E_{d,\chi^2}^+}{\partial p_m \partial p_l} = 
\begin{pmatrix}  2 G_{lm}^{-1} &  -4 \sum_{kno} G_{l k}^{-1} \frac{\partial G_kn}{\partial p_m} G_{n o}^{-1} h_o \\[0.6em]
	-4 \sum_{kno} G_{m k}^{-1} \frac{\partial G_kn}{\partial p_l} G_{n o}^{-1} h_o  & 
	6 \sum_{knowrqp} h_k G_{k o}^{-1} \frac{\partial G_ow}{\partial p_l} G_{w r}^{-1} \frac{\partial G_rp}{\partial p_m}G_{p q}^{-1} h_q 
\end{pmatrix}
$$ 
Since in $E_{d,\chi^2}^+$ is a quadratic formula and only the fit function is varied, its expectation value can
not be lower than the minimum taken at $F^\text{opt}_{d}(x)$ by construction. In most cases the second degree
Taylor polynomial approximation in eq.~\eqref{eq:ExpectedLossFunction} in the $p_m$ variables is enough as it
is quadratic in the $h_i$ and the dependence on the $G^{-1}_{ik}$ is usually weak, because $G_{ik}$ depends on
the sampling in $x$ and does not affect the $\mathbb{E}(y|x)$ conditional expectation value in the large sample
size limit. Therefore for large sample sizes and approximately error-free $G_{ik}$ matrices $\frac{\partial^2 E_{d,\chi^2}^+}{\partial p_m \partial p_l}$ can be approximated with the its upper left submatrix, that
represents the derivatives of $h_l$ and $h_m$. In this approximation $E_{d,\chi^2}^+$ is a $\chi^2$ variable with $d$ degrees of freedom, given that the covariance matrix $\cov{h_i, h_k}$ is non-singular. 

In the full calculation of the standard deviaton of $E_{d,\chi^2}^+$ one should include the variable $\avg{y^2}$
and its correlation to the other input parameters $p_m$. Nevertheless, the $\avg{y^2}$ term disappears when a
loss function difference is calculated for two different fit functions, and it can be shown that the variance
 of the loss function differences can be calculated solely from the variance of the individual $E_{d,\chi^2}^+$
 terms, neglecting the contributions from $\avg{y^2}$. This type of variance is twice the square of the
 previously described bias term in the expectation value in eq.~\eqref{eq:ExpectedLossFunction}, as it would be
 exptected for a $\chi^2$ distribution:

$$ \sigma^2_{d,E^+} = 2 \left( \frac{1}{2}\sum_{lm} \frac{\partial^2 E_{d,\chi^2}^+}{\partial p_m \partial p_l} \cov{p_m, p_l} \right)^2 \,.$$
Similar to the behaviour of $\chi^2$ differences, the expectation value of the $E_{d2,\chi^2}^+ - E_{d1,\chi^2}^+$   difference is the difference of their expectation values. The same can be said about the variance of $E_{d2,\chi^2}^+ - E_{d1,\chi^2}^+$, as it can be calcuated from the square root of the individual $E_{d2,\chi^2}^+$ variations:
$$ \sigma_{d_1,d_2,\Delta E^+} =| \sigma_{d_2,E^+} - \sigma_{d_1,E^+}| \,.$$
This addition rule makes it extremely simple to find the optimal degrees of freedom $d$, as there is no need to
calculate covariance matrices belonging to the different fit functions. This makes the significance of a
feature not only relatively true, but absolute. To optimize the degree of freedom $d$ with a certain
significance level, it is enough to minimize $\mathbb{E}(E_{d,\chi^2}^+ )+ s\sigma_{d,E^+}$, where $s$ can be
arbitrarily chosen. Due to the characteristics of the $\chi^2$ distribution, optimizing for the minimal expected $E_{d,\chi^2}^+ $ results in selecting features that are at least $50\%$ significant.

\begin{figure}[]
\centering
\subfloat[dependence of the expected loss function on the degrees of freedom]{\includegraphics[width=0.42\linewidth]{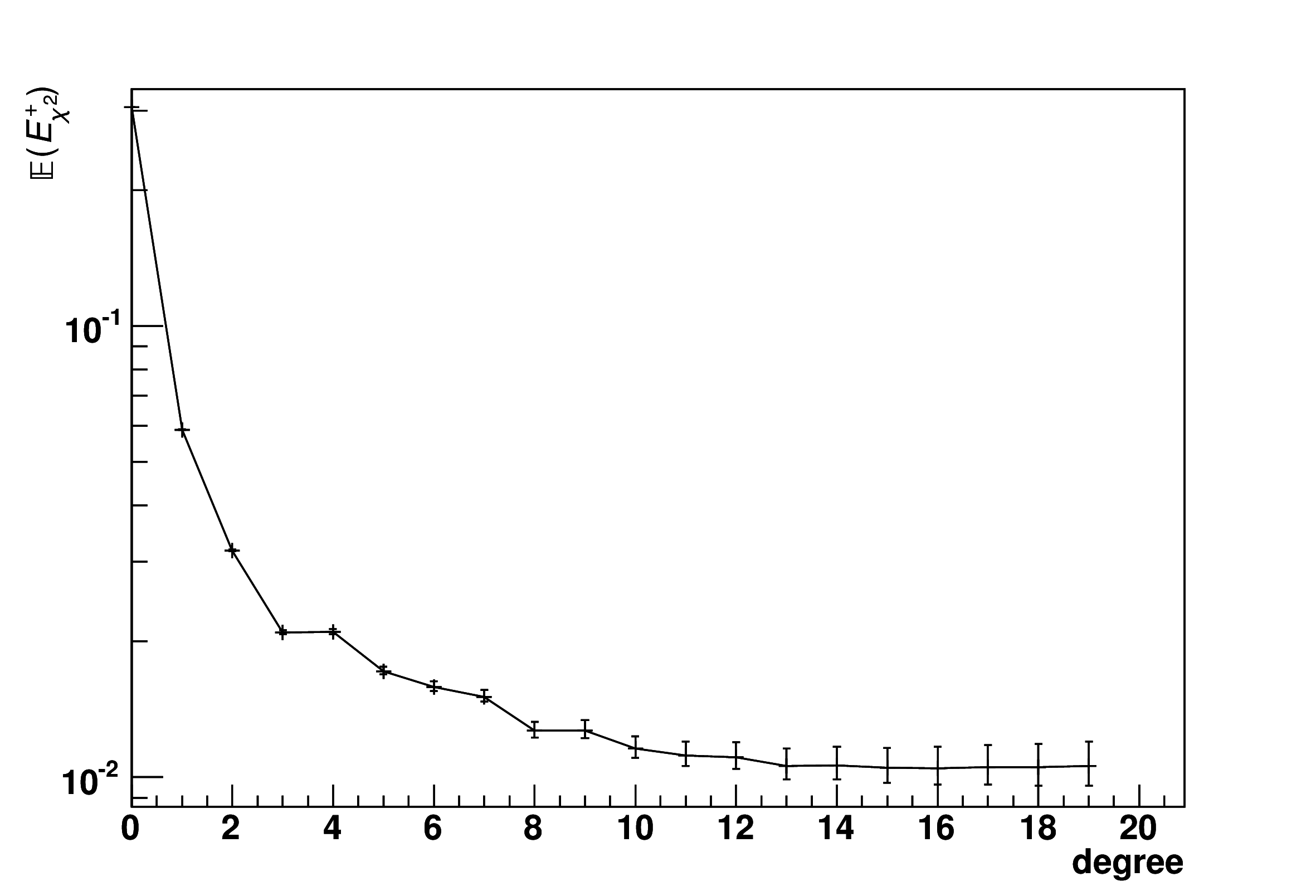}\label{subfig:ExpectedLoss}}%
\subfloat[regression with a single 16 degree polynomial]{\includegraphics[width=0.42\linewidth]{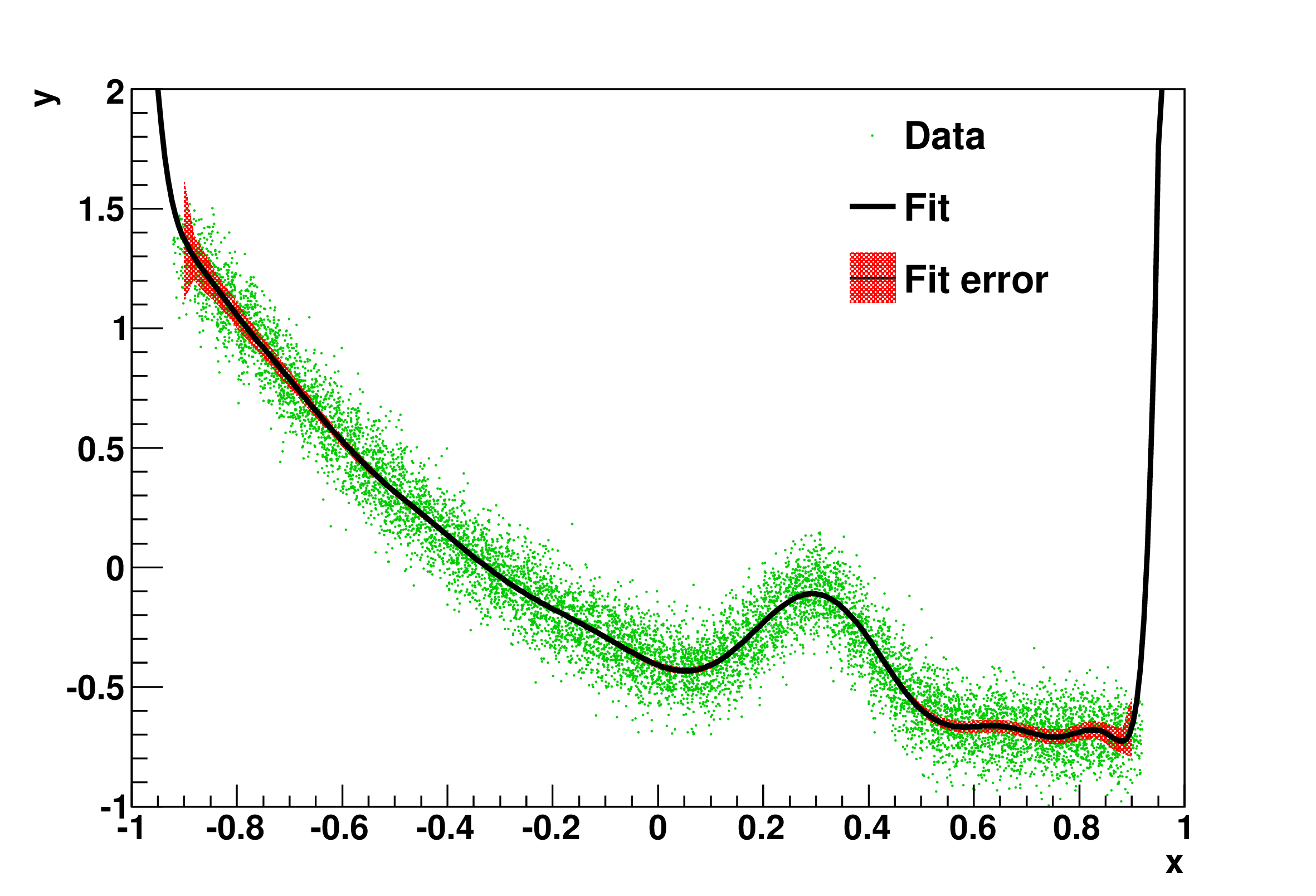}\label{subfig:16degreeOptimal}}\\
\subfloat[regression with constants]{\includegraphics[width=0.42\linewidth]{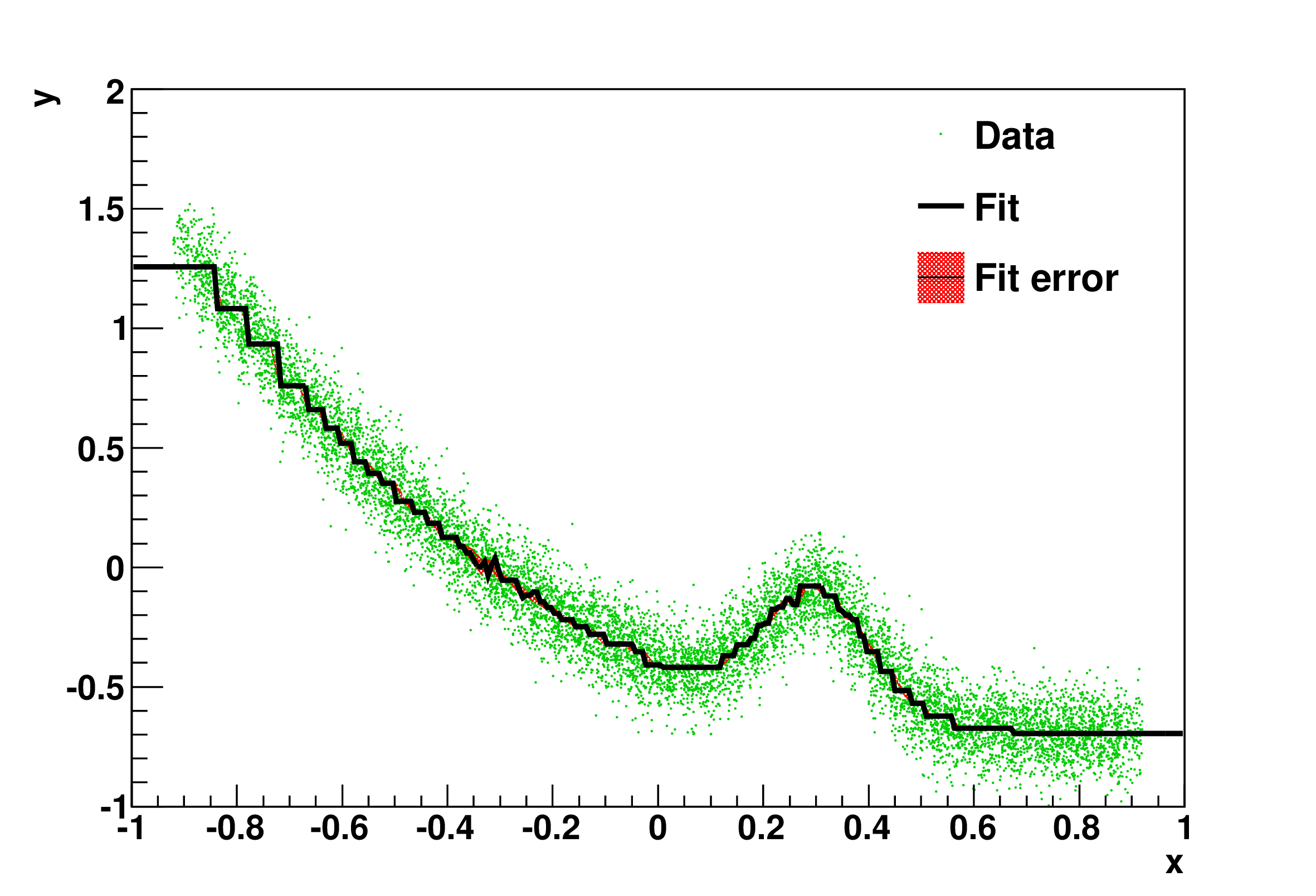}\label{subfig:ConstantsRobot}}
\subfloat[regression with linear functions]{\includegraphics[width=0.42\linewidth]{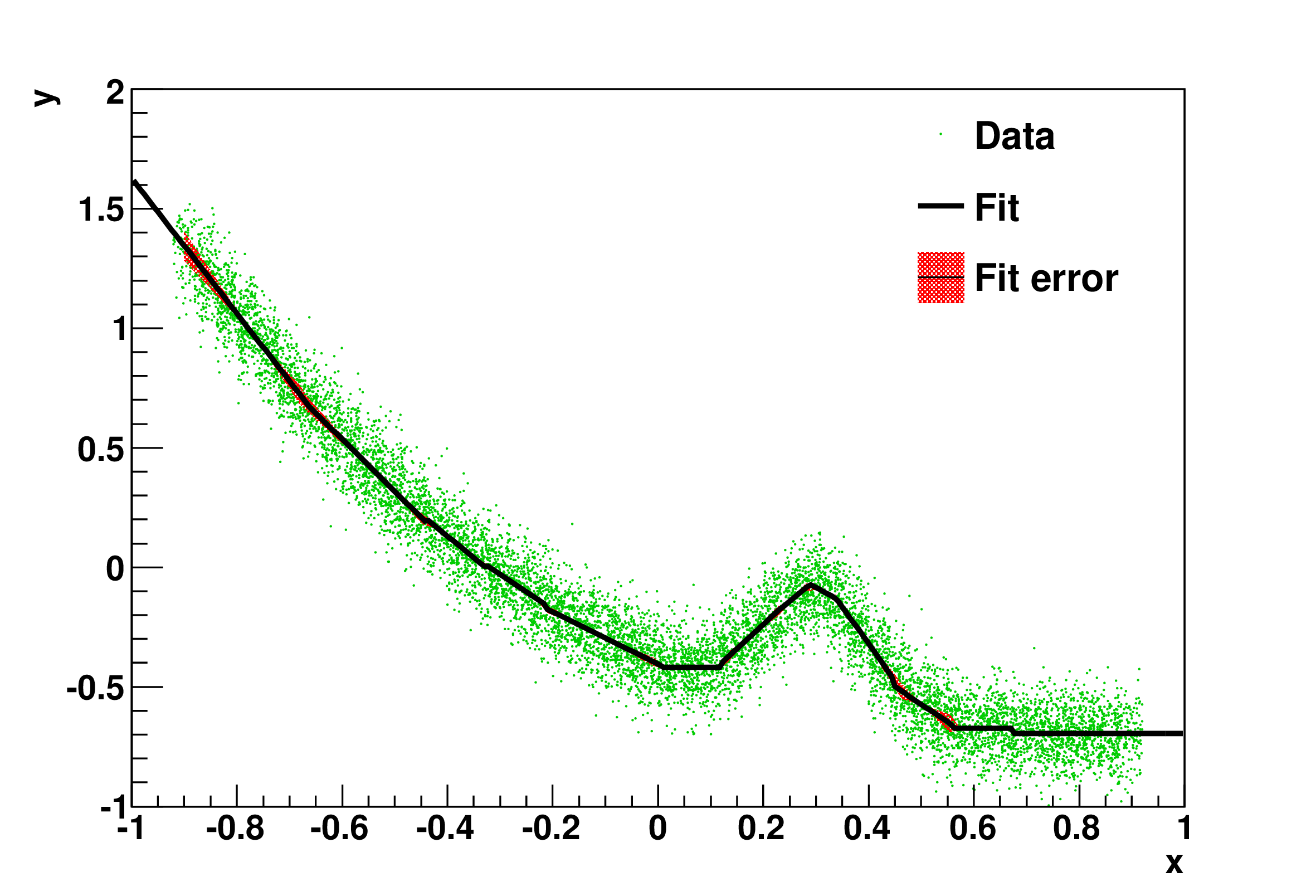}\label{subfig:LinearRobot}}
\caption[]{Example univariate regression. The training sample was a uniform distribution in $x\in(-0.9, 0.9)$ and a Guassian smearing in the y direction for every $x$.
The expected loss function in fig.~\ref{subfig:ExpectedLoss} has a minimum for the 16 degree polynomial. The bottom of the 
error bar represents the $E_{d,\chi^2} $ loss function evaluated with the optimal polynomial, the data points are the $\mathbb{E}{(E_{d,\chi^2}^+)}$ expectation values, while the upper error bars show the one standard deviations of $E_{d,\chi^2}^+$ above the expectation value. Figure~\ref{subfig:16degreeOptimal} shows the 16 degree optimal polynomial with very large uncertainties at the two 
edge of the sample. Figure~\ref{subfig:ConstantsRobot} and~\ref{subfig:LinearRobot} shows regressing with a splitting procedure described in the bulk of the text, with allowed maximal polynomial degrees $n_\text{max} = 0$ and $1$ accordingly.  Both of these regressions functions are within one sigma of the conditional $\prob{y|x}$ distribution and approximates the $\mathbb{E}(y|x)$ conditional expectation value without picking up statistical noise. 
 }\label{fig:UnivariateRobot}
\end{figure}

\section{Determining the right polynomial order}

Selecting the right modelling function can be understood as a repeated hypothesis testing. One must decide 
\emph{a priori} about the null hypothesis and the series of hypotheses to test. In case of polynomial fitting,
the ordering seems trivial, going from a constant to higher polynomial orders. However, with multivariate
input $x_\mu$, a polynomial can be defined with a different degree belonging to each $\mu$ index. In this case,
one may still choose a common order, as the optimizing method described in the previous section allows the
comparison of fit functions differing in multiple degrees of freedom. A common polynomial order is special in the
sense that it is closed under the group of rotations and translations, treating the different $\mu$ directions
equally. 

However, it is not possible to compare the loss functions of the infinite many polynomials. Not only because it is
infeasible, but alse because one must stop
before the numerical errors overcome the estimated uncertainties in $E^+_{\chi^2}$. It is not straightforward
to estimate these numerical errors, but as a guideline it can be said that it increases with both the
number of polynomial degrees and the number of input dimensions in the multivariate case. The polynomial
degree necessitate an exponentiation on the input parameters, which both appears in the training  and 
in the function evaluation phase. A double precision number can be thought of as a sixteen digit decimal
number, and though its $n^\text{th}$ exponential is expected to have a relative error of only $n\cdot10^{-16}$,
the numerous subtractions and multiplications needed for the linear equation solver can easily blow this up.
In the univariate case, it numerical errors seems to become significant at the polynomial order around 20 for 
double precision and $24...30$ for 128bit long double precision. In the case of $d$ input dimensions the size 
of the $G$ matrix grows rapidly with the degrees. It is because the polynomial coefficients of the $F(x)$ fit 
function are $d$-degree symmetric tensors, with $\binom{n+d-1}{n}$ free parameters, and all the free parameters
in $F(x)$ contribute to the size of the $G$ matrix. For 20 input dimensions a 3 degree polynomial has
nearly 2000 free parameters, resulting in $G$ matrix size of $2000\times2000$, and although solving 
a linear equation with this only takes a few seconds on a modern-day computer, this also means that 
more than a million instructions are needed to express each unknown of the $F(x)$ polynomial, 
resulting in large numerical errors. 

\newsavebox{\tempbox}
\begin{figure}[]
\centering
\sbox{\tempbox}{\includegraphics[width=.47\linewidth]{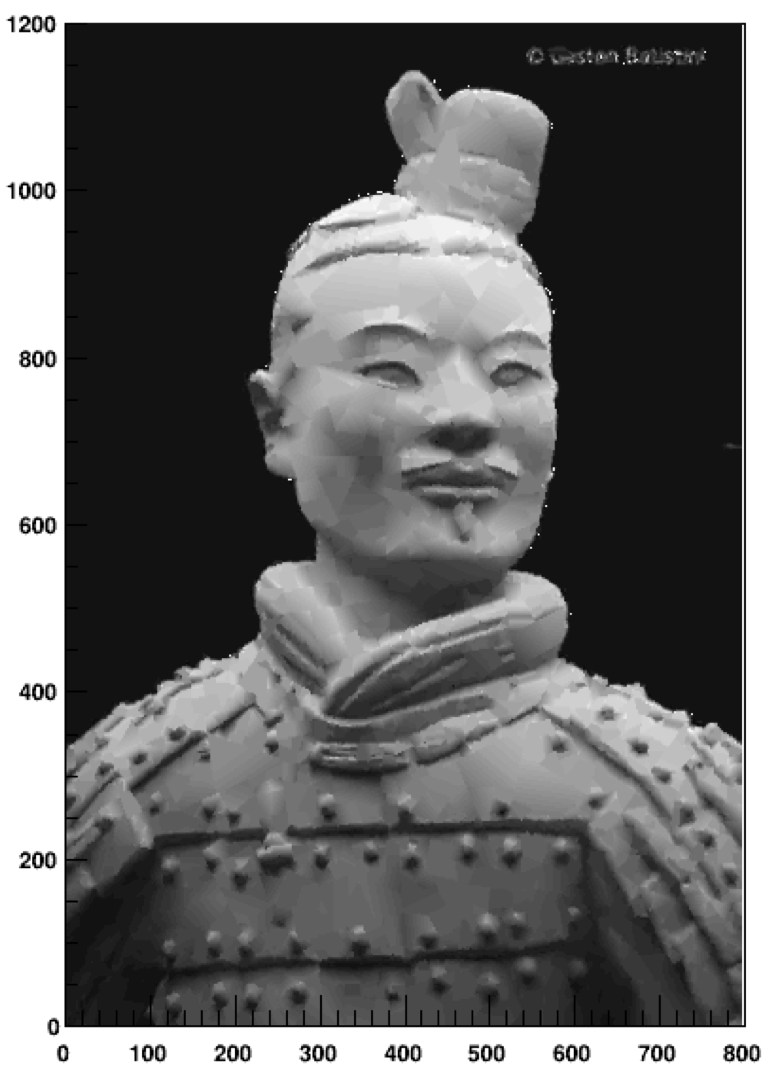}}%
\subfloat[original]{\label{fig:originalArcher}%
\vbox to \ht\tempbox{\vfill\hbox to \wd\tempbox{%
\includegraphics[width=0.40\linewidth]{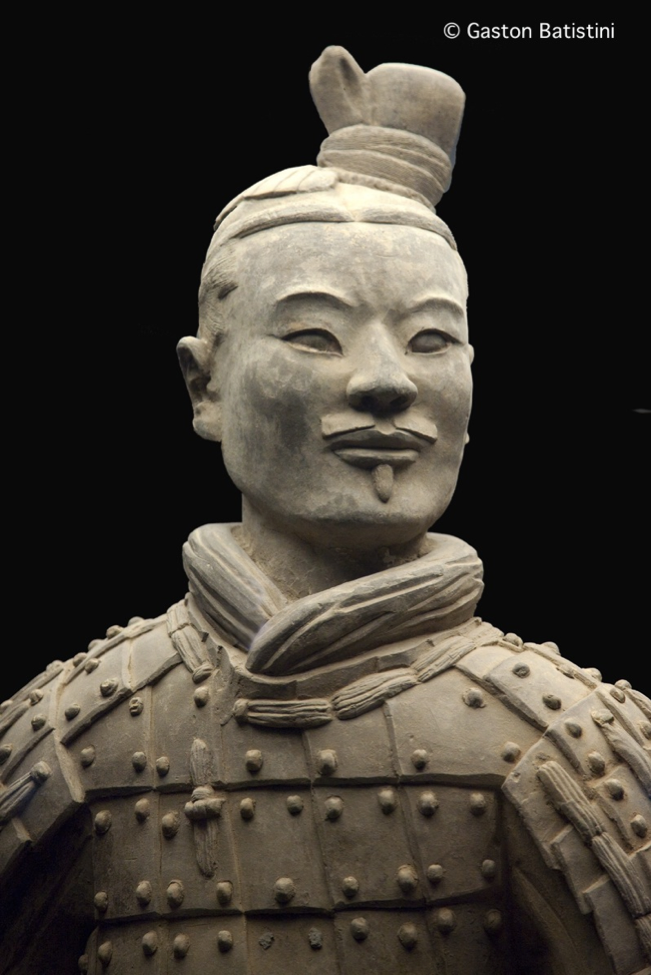}%
}\vfill}%
}%
\subfloat[remodelled]{\label{fig:remodelledArcher}\usebox{\tempbox}}%
\caption[two archer]{
Figure~\subref{fig:originalArcher} shows a photo of a Chinese terra-cotta soldier, whose intensity map was remodelled in
 fig.~\subref{fig:remodelledArcher} in small patches with polynomials. The photo was treated as a random sample, namely that prior knowledge of the uncertainty of the pixel intensities were neglected. The region boundaries were determined with algorithm 
described in the bulk of the text, based on the principal axes of the input distribution. The regions were modelled with a two 
dimensional linear polynomial, with a second layer of univariate regression upon it, to simulate a sigmoid-like behaviour. Over- and
undershoots from the displayable range were rounded to the maximum and minimum intensities accordingly. 
A photo instead of a real 3D distribution was chosen here in order to demonstrate that this simple fitting and splitting method can
detect the significant features in the data while smoothing the small details, as one would require for a regression method on a statistical sample.  
}\label{fig:MultivariateRobot}%
\end{figure}

\section{Minimizing numerical errors}

To overcome the problem of the high polynomial degrees and the large matrices, one can split the sample into 
many smaller phase spaces, which may require smaller polynomial degrees. For this one must 
decide on a maximal polynomial degree $n_\text{max}$ which is allowed in regression, and
$n_\text{max} + k$, $k>1$ which up to expected loss function is scanned. If the optimal degree
of the expected loss function appears to be larger than $n_\text{max}$, one can apply a predefined
algorithm that splits the input phase space. One such algorithm for the univariate case simply splits 
the input phase space at the $x$ mean, as demonstrated on fig.~\ref{fig:UnivariateRobot}. 
This requires practically no additional computation, since the $\avg{x}$ was already calculated for 
the regression. The splitting and fitting can be repeated until the full sample is regressed.
A possible extension of this approach to the multivariate case finds the multivariate 
mean $\avg{x_\mu}$ first, then splits the sample at this point parallel to the principal axis, the 
eigenvector with the largest eigenvalue of the $\avg{  x_\mu x_\nu } - \avg{x_\mu}\avg{x_\nu}$ matrix; 
see fig.~\ref{fig:MultivariateRobot}. These methods have the advantage that they place the cut
boundaries within the distribution, so the regression on these phase spaces
are less likely to produce degenerate solutions. A seemingly more optimal splitting method would be 
finding the place where the fitted polynomial with $n_\text{max} +1 $ degrees have the largest
derivative, since this is a hard place to model with an $n_\text{max}$ degree polynomial.
However, it is non-trivial to define and find this boundary in the multivariate case, and
this boundary typically appear nearby the tails of the $x$ distribution, where the fitted function has the 
largest uncertainty.

\section{Conclusions}

The presented method is capable of modelling multivariate statistical data with polynomials by detecting 
the significant features in the data. It is a fast and robust method, as most calculations are computationally
very simple and it does not require numerical optimisation. Similarly to the statistical bootstrap 
method, the uncertainty of the regression function can be determined from the training sample, 
but in this case analytically. In combination with a phase space splitting method, it can be extended to fit
very complex data, still maintaining numerical stability. 

\nocite{2012arXiv1203.5647K}
\nocite{KovesarkiPhDthesis}
\nocite{KovesarkiACAT2013}
\nocite{Metzger2010}
\nocite{HankelMatrix}
\nocite{Bishop:2006:PRM:1162264}
\nocite{Ripley:1996PRandNN}
\nocite{2013ApJ...764..167S}

\section*{References}
\bibliographystyle{BibTeX/iopart-num/iopart-num}
\bibliography{references.bib}

\end{document}